%% file: main.tex
\documentclass[lettersize,journal]{IEEEtran}
\usepackage{amssymb, amsthm, amsmath, amsfonts}
\usepackage{algpseudocode}
\usepackage{algorithm}
\usepackage{tikz}
\usepackage{array}
\usepackage[caption=false,font=normalsize,labelfont=sf,textfont=sf]{subfig}
\usepackage{textcomp}
\usepackage{stfloats}
\usepackage{url}
\usepackage{verbatim}
\usepackage{graphicx}
\usepackage{cite}
\usepackage{nameref,hyperref}
\usepackage{stfloats}
\usepackage{multirow}

\usetikzlibrary{decorations.pathreplacing}
\usetikzlibrary{calc}
\theoremstyle{definition}

\DeclareMathOperator*{\argmax}{arg\,max}

\makeatletter
\newcommand{\thickline}{%
	\noalign {\ifnum 0=`}\fi \hrule height 1pt 
	\futurelet \reserved@a \@xhline 
	}
\makeatother

\begin{document}
\title{\textbf{DOST} - Domain Obedient Self-supervised Training for Multi Label Classification with Noisy Labels}
\author{Soumadeep Saha, Utpal Garain, Arijit Ukil, Arpan Pal, Sundeep Khandelwal \\
\vspace{2em}
\color{blue} \copyright 2023 IEEE. Personal use of this material is permitted. Permission from IEEE must be obtained for all other uses, in any current or future media, including reprinting/republishing this material for advertising or promotional purposes, creating new collective works, for resale or redistribution to servers or lists, or reuse of any copyrighted component of this work in other works.
}
\maketitle

\begin{abstract}
\input{chapters/abstract.latex}
\end{abstract}

\begin{IEEEkeywords}
Domain knowledge, logical constraints, noisy labels, multi-label classification, deep learning
\end{IEEEkeywords}

\input{chapters/intro.latex}
\input{chapters/related.latex}
\input{chapters/background.latex}
\input{chapters/method.latex}
\input{chapters/experiments.latex}
\input{chapters/results.latex}
\input{chapters/conclusions.latex}

\newpage
\input{chapters/appendix.latex}
\end{document}

%% file: chapters/abstract.latex
The enormous demand for annotated data brought forth by deep learning techniques has been accompanied by the problem of annotation noise.
Although this issue has been widely discussed in machine learning literature, it has been relatively unexplored in the context of ``multi-label classification'' (MLC) tasks which feature more complicated kinds of noise.
Additionally, when the domain in question has certain logical constraints, noisy annotations often exacerbate their violations, making such a system unacceptable to an expert.
This paper studies the effect of label noise on domain rule violation incidents in the MLC task, and incorporates domain rules into our learning algorithm to mitigate the effect of noise.
We propose the \textbf{D}omain \textbf{O}bedient \textbf{S}elf-supervised \textbf{T}raining (\textbf{DOST}) paradigm which not only makes deep learning models more aligned to domain rules, but also improves learning performance in key metrics and minimizes the effect of annotation noise.
This novel approach uses domain guidance to detect offending annotations and deter rule-violating predictions in a self-supervised manner, thus making it more ``data efficient'' and domain compliant.
Empirical studies, performed over two large scale multi-label classification datasets, demonstrate that our method results in improvement across the board, and often entirely counteracts the effect of noise.

%% file: chapters/intro.latex
\section{Introduction}
\label{sec:intro}
\IEEEPARstart{S}{upervised} deep learning algorithms have been massively successful in tackling problems in diverse domains.
However, their Achilles heel seems to be the copious amounts of annotated data required, which often leads to annotation techniques which are somewhat unreliable and introduce \textbf{label noise}\footnote{We use annotation and label interchangeably}.
The situation gets worse when the applications require domain specific expertise as labeling the data becomes an extremely resource intensive endeavour (\cite{Esteva2017}).
This results in smaller datasets \cite{notEnoughData} which still might contain significant quantities of noise.
Experts often disagree on a label (inter-observer variability), and uncertainties which might arise (annotator error) are often ignored.
Often, annotations are auto generated with algorithms, or crowdsourced with the help of novices, and thus annotations are of lower quality (\cite{Kuznetsova2020, dataBad}).
Considering the fact that deep learning models are prone to overfitting, and have even been shown to fit randomly generated labels \cite{overfit}, label noise often acts as a major hurdle (\cite{noisyFace, Kuznetsova2020}) in developing reliable deep learning systems.

Multi label classification (MLC), where a subset of target classes must be predicted from a single data instance, is a standard problem in several domains like computer vision, natural language processing, computational diagnostics \cite{mlCV, mlCV2, mlNLP, mlMED}, etc.
However, despite the widespread usage of deep learning based MLC techniques, the problem posed by noisy labels in MLC has not received adequate research attention.
It is also a significantly harder problem compared to multi-class classification, requiring an arbitrary number of classes to be predicted, thus increasing the number of distinct annotations from $|A|$ to $\sim 2^{|A|}$ for target classes in $A$, making errors more likely. 
Furthermore, unlike the single label setting, where a mislabeled instance has exactly two classes flipped, more complex errors are possible in the multi label setting.
For instance a subset or superset of the target classes may have been annotated, which still retain valuable information, and thus treating these incidents in the same light as completely erroneous entries is undesirable.

In this paper, we focus on a specific kind of annotation error - those that \textbf{contradict the domain rules}. 
For example consider an image classification system with the rule that an image shouldn't be marked as the classes ``bird'' and ``mammal'' simultaneously, or a medical diagnostic tool that shouldn't mark a patient as ``hypertensive'' and ``hypo-tensive'' at the same time.
Requirements of these kind are seen across many applications, and deep-learning algorithms do not learn to abide by them natively.
This lack of domain rule compliance, which is compounded by noisy annotations, hinders adoption of machine learning systems often in domains with the most pressing need, such as digital theraputics, law, etc.
This problem is exacerbated when annotations are inconsistent with domain rules, and the models trained on these datasets tend to produce these inconsistencies in abundance at inference time. 
Errors of this kind are much more deleterious than simple mislabelling for widespread adoption of these systems, as it appears to be ``illogical'' from a domain expert perspective.

One naive solution to deal with this would be to simply remove these contradictory instances from the dataset based on the available domain rules, however, as we are often starved for high quality data to begin with, this should be the last resort.
The question we seek to answer in this paper is ``Can we do better?'' i.e. can we make use of erroneously annotated data in conjunction with the domain rules to produce models which are more coherent with the domain rules (does not make inferences contrary to domain rules) and perform at par or better than models trained on the reduced dataset?

Our investigations in this regard first lead us to measure the effect of annotation noise on the performance of models, both in terms of standard metrics as well as adherence to domain rules.
Further, we propose a novel Domain Obedient Self-supervised Training (DOST) paradigm that produces inferences which are more coherent with the rule set and outperforms models trained on the reduced dataset.
Our approach is more data efficient, and can use domain rule violating instances to mitigate the effects of label noise to a large extent.
This approach also imbibes the model with semantic information from the domain, which leads to improved performance.

In the coming sections we provide a concise discussion on related works that forays into this field, followed by a formal description of the problem and our proposed solution. We then present experimental results on two large scale MLC datasets PASCAL VOC \cite{Everingham15} and PhysioNet CiNC 2020 \cite{cmpaper} followed by a few concluding remarks.

%% file: chapters/related.latex
\section{Related Work}
\label{sec:related}
Label noise has been a prevalent problem plaguing deep learning systems, and several approaches have been proposed to deal with them \cite{noiseSurvey, noisyMedical}.
Some of these approaches include label cleaning and pre-processing \cite{Vo2015EffectiveTO}, modifying network architectures \cite{NIPS2017_e6af401c}, introducing robust loss functions and regularisation \cite{Ghosh_Kumar_Sastry_2017, NEURIPS2018_ad554d8c}.
Compared to the rich discourse in the single label setting, efforts in the multi label setting have been rather sparse.
The standard approach in MLC is to decompose it into a set of independent binary classification problems.
Ironically, recent successes in MLC rely on exploiting label correlations \cite{Wang_He_Li_Long_Zhou_Ma_Wen_2020, 8953276}, which makes the issue worse when noise is involved.
To counter this Zhao and Gomez \cite{Zhao2021EvaluatingMC} introduced a loss function to learn robust embeddings.

This problem is often framed through the lens of disambiguation i.e. given a dataset containing noisy annotations we first try to recover ground-truth labeling information from candidate labels before employing learning algorithms.
Xie and Huang \cite{Xie_Huang_2018}, Fang and Zhang \cite{Fang_Zhang_2019} attempted introducing labeling confidences, whereas Li-juan Sun and Jin\cite{Sun_Feng_Wang_Lang_Jin_2019} employed a low-rank and sparse decomposition scheme.
Garcia et al. \cite{GARCIA201614} proposed a meta-learning system recommending the expected performance of noise filters in noisy data identification task.
Xie and Huang \cite{9354590} proposed a framework for MLC that simultaneously identifies noisy labels.

Our work forays into domain knowledge augmentation, and in particular deals with logical constraints arising out of domain information.
Logically constrained MLC has been widely explored, and tends to outperform standard MLC systems on certain tasks. 
Giunchiglia and Lukasiewicz \cite{NEURIPS2020_6dd4e10e} proposed a method for tackling the related hierarchical MLC problem by appending a constraint layer to their models which is trained to exploit hierarchies in the annotations.
This was shown to be an improvement over standard techniques which involve post-processing, and does not pass the constraining information to the underlying model.
Melacci et al. \cite{9661418} trained MLCs with an augmented loss function which expresses logic constraints as a polynomial of predicate probabilities and showed that they help alleviate adversarial attacks.

To the best of our knowledge the problem of logically constrained MLC has not been studied in a noisy label setting, where such constraint violations would be rampant.
Our proposed method \textbf{DOST}, utilizes the opportunity provided by these logical constraints to identify noisy labels, and simultaneously improve performance and domain coherence. 

%% file: chapters/background.latex
\section{Background}
\label{sec:background}

\subsection*{Notation}
We start with a dataset $\mathcal{D} = \{(x_i, \hat{y}_i)| \: \forall i \in \{1, \ldots, N\}\}$ where $x_i \in \Lambda$ are the data instances and $\hat{y}_i$ are the corresponding (potentially erroneous) labels.
We have a set of target classes $A = \{ a_1, a_2, \ldots, a_p \}$, and we have $\hat{y}_i \subseteq A \; \forall i \in \{1, \ldots, N\}$.
Let $a_j$ also denote a unary predicate, and $a_j(x_i)$ is True, if $x_i$ is an instance of $a_j$ else False.
So, the ordered pair $(x_i, \hat{y}_i)$ induces the formula 
\begin{equation}
	\begin{split}
	\mathrm{F}(x_i,\hat{y}_i) = a_{\beta_1}(x_i) \wedge a_{\beta_2}(x_i) \wedge \ldots a_{\beta_t}(x_i) \wedge \\
		\neg a_{\gamma_{t+1}}(x_i) \wedge \neg a_{\gamma_{t+2}}(x_i) \wedge \ldots \neg a_{\gamma_p}(x_i)
	\end{split}
\end{equation}

Where, $\forall j$ $a_{\beta_j}$ are in $\hat{y}_i$ and $a_{\gamma_j}$ are not in $\hat{y}_i$.

We also have a set of domain rules $\mathrm{R} = \{r_1, r_2, \ldots \}$, where each $r_k$ takes the form
\begin{equation}
	r_k = \forall x, \: \phi_k \rightarrow \psi_k
\end{equation}
Where $\phi_k, \psi_k$ are first-order logic formulas built from the predicates in $A$ and the logical connectives $\{\wedge, \vee, \neg\}$.

We call an annotation $y_i$ of $x_i$ \textbf{contradictory} or a \textbf{contradiction} (denoted $C(y_i)$) with the rules $\mathrm{R}$, iff
\begin{equation}
	\exists \: r \in \mathrm{R}, \text{such that, } r \Rightarrow \neg \mathrm{F}(x_i,y_i)
\end{equation}
To measure the amount of contradiction in a set of annotations $Y = \{y_i|i \in \{1\ldots n\}\}$ we use the metric $C(Y)$ which is defined by the following equation.
\begin{equation}
	\label{eq:contradiction-metric}
	C(Y) = \frac{1}{|Y|}\sum_{i=1}^{n} \text{Number of rule violations in }y_i
\end{equation}

With the domain rules $\mathrm{R}$ we create two auxiliary datasets, $\mathcal{D}_C$ and $\hat{\mathcal{D}} = \mathcal{D} - \mathcal{D}_C$, where $\mathcal{D}_C$ is the subset of the dataset with contradictory annotations, i.e. $(x_i,\hat{y}_i) \in \mathcal{D}_C \iff y_i$ is a contradictory annotation. Let $|\mathcal{D}_C| = N_C$ and $p_c = \frac{N_C}{N}$ represent the proportion of contradictory samples.

\subsection*{Problem Setup}
Given dataset $\mathcal{D}$ and a rule set $\mathrm{R}$ we want $f_{\nu}$
\begin{equation}
\begin{split}
	f_{\nu} : \Lambda &\rightarrow [0,1]^p\\
	[f_{\nu}(x)]_k &\approx P\big(a_k(x)\big)
\end{split}
\end{equation}

The MLC prediction $y$ is then given by
\begin{equation}
	y = g_{\nu,\mu}(x) = \{a_k | [f_{\nu}(x)]_k \geq \mu, \: \forall a_k \in A \}
\end{equation}
for some constant $\mu \in [0, 1]$. Ideally, we should have

\begin{align}
	P\Big(a_k \in g_{\nu,\mu}(x)\Big|& a_k(x)\Big) \rightarrow 1\\
	P\Big(a_k \in g_{\nu,\mu}(x)\Big|& \neg a_k(x)\Big) \rightarrow 0\\
	P\Big(C\big(g_{\nu,\mu}(x)\big)\Big)& \rightarrow 0
\end{align}
Or, simply put, we want to have an accurate multi label classifier which avoids contradictory annotations.

 In this paper we only consider rules of the form $\forall x,\: a_i(x) \Rightarrow \neg \big(a_{\beta_1}(x) \vee a_{\beta_2}(x) \vee \ldots a_{\beta_t}(x)\big)$ for $a_{\beta_l} \in A$ ($\beta_l \neq i$) in order to simplify creation of $\mathrm{R}$.
 However, this is general enough to capture most rule sets in use in a MLC context.

%% file: chapters/method.latex
\section{\textbf{DOST} - A Domain Obedient Self-supervised Multi Label Classifier}
\label{sec:method}

In this section we outline \textbf{DOST}, the domain obedient self-supervised training algorithm in order to find $g_{\nu,\mu}$ such that $P(C(g_{\nu,\mu}))$ is minimized alongside the classification targets.
Given $\mathcal{D}$ and $\mathrm{R}$ we first construct $\hat{\mathcal{D}}$ and $\mathcal{D}_C$ as mentioned in the preceding section.
Additionally, we create a rule matrix $\rho \in \mathbb{R}^{p\times p}$, which is defined as
\begin{equation}
	\rho_{ij} = \begin{cases}
		1 & \text{if } \exists\; r \in \mathrm{R}, \text{s.t. } r \Rightarrow \;\forall \;x\in\Lambda,\; a_i(x) \Rightarrow \neg a_j(x) \\
		0 & \text{ otherwise.}
	\end{cases}
\end{equation}

\begin{figure}[tpb]
	\centering
	\caption{\textbf{Schematic diagram of DOST algorithm}}
	\includegraphics[width=3in]{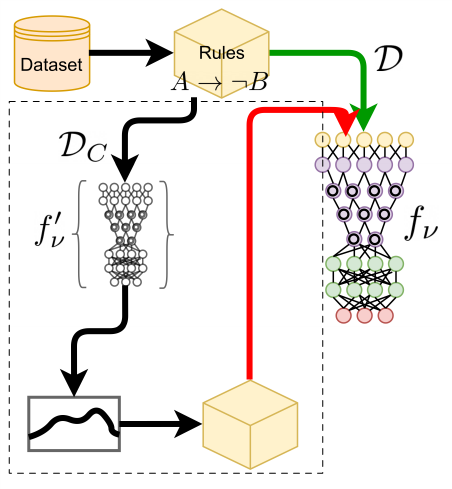}
	\label{fig:dost}
\end{figure}

With $\hat{\mathcal{D}}$, $\mathcal{D}_C$ and $\rho$, we train a deep neural network (with sigmoid outputs) on $\hat{\mathcal{D}}$ as usual (with Binary Cross-entropy loss); and for samples in $\mathcal{D}_C$ we use the target
\begin{equation}
	\begin{split}
		\delta_C(x_i) &= \rho\Big[\argmax\big(f'_{\mu}(x_i)\big)\Big]\\
		\mathcal{L}_C (x_i) &= -\zeta\cdot \sum_{j=1}^p \Big[\delta_C(x_i)_j\cdot \log\big(1 - f_{\mu}(x_i)_j\big) \Big]
	\end{split}
	\label{eq:loss}
\end{equation}
Where $\zeta$ is a constant (hyper-parameter), and $f_{\mu}$ represents the deep network with parameters $\mu$ ($f'_{\mu}$ is a slightly older copy of the network). Thus in effect, we find potential classes which result in rule violations based on the model's best guesses, and we use those as negative examples while training on $\mathcal{D}$. Since $\delta_C(x_i)$ is not a function of $\mu$, $\mathcal{L}_C$ is differentiable and can be used with any standard gradient based training method. The algorithm is outlined in Algorithm \ref{alg:dost}, and a schematic diagram is given in \ref{fig:dost}.

\begin{algorithm}
\caption{\textbf{: DOST}}
\label{alg:dost}
\begin{algorithmic}[1]
\While{not StopCondition}
	\State Get a batch of samples from $\mathcal{D}$
	\State $l \leftarrow 0$
	\For{$(x_i,\hat{y}_i)$ in batch}
		\If{$(x_i,y_i) \in \hat{\mathcal{D}}$}
			\State $l \mathrel{+}= \mathcal{L}\Big(\hat{y}_i, f_{\mu}(x_i)\Big)$ \Comment{Classification loss}
		\Else
			\State $l \mathrel{+}= \mathcal{L}_C (x_i)$ \Comment{Eq. \ref{eq:loss}}
		\EndIf
	\EndFor
	\State $\text{Back-propagate}(l)$
	\State Update model $\mu$
	\State Every k steps $f' \leftarrow f$
	\If{$\ldots$} \Comment{Check if model converged}
		\State StopCondition $\leftarrow$ True
	\EndIf
\EndWhile
\end{algorithmic}
\end{algorithm}

%% file: chapters/experiments.latex
\section{Experiments}
\label{sec:experiments}

We perform our experiments on two large scale multi-label datasets, PASCAL VOC \cite{Everingham15} and PhysioNet 2020/21 \cite{cmpaper}.
PASCAL VOC is a commonly used computer vision MLC benchmark, requiring us to pick the subset of classes present in an image from a set of 20 possible classes.
PASCAL VOC does not come with a set of rules precluding certain class combinations, however for demonstration purposes, we constructed a rule set based on classes which never appear together in the training set.

The PhysioNet dataset is a conglomeration of several large scale 12 lead Electrocardiogram (ECG) datasets annotated with 24 unique cardiovascular features.
For this dataset we created a rule set in collaboration with practising cardiologists based on widely known facts about human cardiovascular physiology.
For, example we know that a patient exhibiting Atrial Fibrillation will not show Normal Sinus Rhythm, thus precluding these two classes from co-occurring in a diagnosis. Further details about the rule set used in each case can be found in the Appendix.

Having created the rule set from co-occurrence of classes in the dataset, PASCAL VOC annotations are obviously not contradictory with the rule set.
In the PhysioNet dataset, we found $\approx 20\%$ of the annotations were contradictory with the rule set.
In both cases to measure effect of noise on performance, we introduce additional noise.

With probability of $p_c$ we introduce a noisy instance, either by sampling from the set of noisy instances or by injecting noise following a geometric distribution.
Following this, from the dataset $\mathcal{D}$ and the rule set $\mathrm{R}$ we create the datasets $\hat{\mathcal{D}}$ and $\mathcal{D}_C$ such that $p_C = \frac{\mathcal{D}_C}{\mathcal{D}}$.
We then use these to measure the effect of noise and the performance of different mitigation techniques.

Our experiments were carried out on a single nVidia RTX A6000 48GB GPU. We report 10 fold cross validation results in all cases.

%% file: chapters/results.latex
\section{Results}
\label{sec:results}

At the very outset, it must be noted that although rare, even without any noisy annotations in the dataset we observe that standard deep learning algorithms still produce contradictory predictions (see Table \ref{tab:cleanCont}).
Our fine-tuned ResNet50 pre-trained \cite{He2015DeepRL} on ImageNet \cite{5206848} for the PASCAL VOC task was comparatively better than the model trained on PhysioNet, perhaps suggesting that improved performance can be seen with pre-training on large datasets.

\begin{table}[htpb]
	\centering
	\caption{Vanilla deep learning model trained on clean dataset produces contradictions (see Eq. \ref{eq:contradiction-metric}).}
	\label{tab:cleanCont}
	\begin{tabular}{|c|c|c|}
		\hline
		Dataset & PhysioNet 2020 & PASCAL VOC \\
		\hline
		Train 		&$4.02\%\pm 1.65$& $0.05\%\pm 0.04$\\
		\textbf{Test}	&$3.57\%\pm 1.32$& $0.11\%\pm 0.09$ \\
		\hline
		
	\end{tabular}
\end{table}

\begin{figure}[htpb]
	\centering
	\caption{\textbf{Effect of noise on performance of deep learning models.}\\
	With increasing amounts of noise model performance continually degrades, whereas contradictions increase.
	The X-axis represents the percentage of training instances containing contradictions.
	(A-E) shows various metrics for PhysioNet dataset, and (F-J) for PASCAL VOC. }
	\includegraphics[width=3.5in]{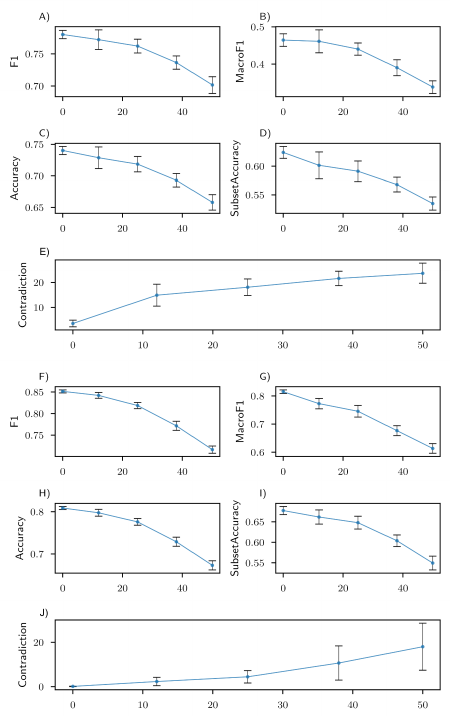}
	\label{fig:noise-effect}
\end{figure}

Our experiments consistently demonstrate that an increase in percentage of noisy annotations made performance worse both in terms of contradictions and model performance in general (see Fig. \ref{fig:noise-effect}).

\begin{figure}[htpb]
	\centering
	\caption{\textbf{DOST paradigm significantly reduces rule violations}\\
	The shaded region in the figure below is a magnified view of the curves close to the X-axis.}
	\includegraphics[width=3.5in]{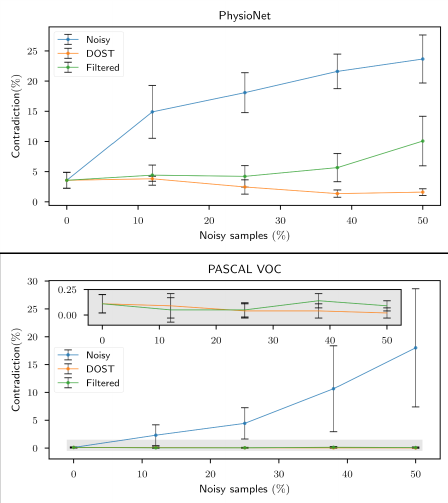}
	\label{fig:noise-dost}
\end{figure}

Following this, we measure the impact of two mitigation strategies, the naive approach of filtering out offending annotations from the dataset and our proposed \textbf{DOST} paradigm, on contradictory outputs (see Fig. \ref{fig:noise-dost}).
Since, we only have the self-supervision loss update on $\mathcal{D}_C$ (see Eq. \ref{eq:loss}, Algorithm \ref{alg:dost}), when $\mathcal{D}_C$ is small the two strategies are very similar, and should have similar outcomes.
We observe that DOST has a significant effect on reducing contradictory predictions, especially at higher levels of noise.

As we noted earlier, the fine-tuned ResNet50 model applied to the PASCAL VOC task has very few contradictory predictions, and this remains unchanged with the filtering strategy (as it was pre-trained on a massive dataset and thus doesn't require as much training data).
We see slight improvements with DOST at higher noise levels, however, the difference is not very significant.
The effect is more pronounced in the model trained from scratch on the PhysioNet dataset, where DOST reduces contradictory outputs to better than baseline levels in most instances (see Table \ref{tab:noise-dost}).
\begin{table}[htpb]
	\centering
	\caption{DOST eliminates contradictory outputs at high noise levels and even outperforms the ideal no noise scenario (data shows $C(Y)$ per hundred instances from the PhysioNet dataset).}
	\label{tab:noise-dost}
	\begin{tabular}{|p{3em}|p{5em}|p{3em}|p{3em}|p{3em}|}
		\thickline
		Noise Level	& Perfect Dataset	& \textbf{DOST}	& Filtered Dataset	& Noisy	Dataset	\\
				& (No noise)		&		&			&		\\
		\thickline
		$12.5\%$ &\multirow{4}{*}{$3.57 \pm 1.32$} & $3.82 \pm 0.41$	& $4.42 \pm 1.67$		& $14.90 \pm 4.38$	\\
		\cline{1-1}\cline{3-5}
		$25\%$     &			& $2.46 \pm 1.19$	& $4.22 \pm 1.78$		& $18.09 \pm 3.31$	\\
		\cline{1-1}\cline{3-5}
		$37.5\%$ & 			& $1.36 \pm 0.61$	& $5.67 \pm 2.33$		& $21.61 \pm 2.86$	\\ 
		\cline{1-1}\cline{3-5}
		$50\%$   &			& $1.61 \pm 0.56$	& $10.07 \pm 4.10$		& $23.66 \pm 3.98$	\\
		\thickline
	\end{tabular}
\end{table}

\begin{figure*}[!t]
	\centering
	\caption{\textbf{DOST paradigm improves performance across several metrics.}\\
	The shaded region above the blue line (performance of model trained on noisy dataset) is the potential room for improvement, given a hypothetical clean dataset, which is usually not available in practice.
	\textbf{DOST} outperforms the naive filtering approach in all cases, and on the PASCAL VOC dataset, almost manages to counteract the effect of noise altogether.
	(A-D) shows various metrics for PASCAL VOC, and (E-H) for PhysioNet. }
	\includegraphics[width=\textwidth]{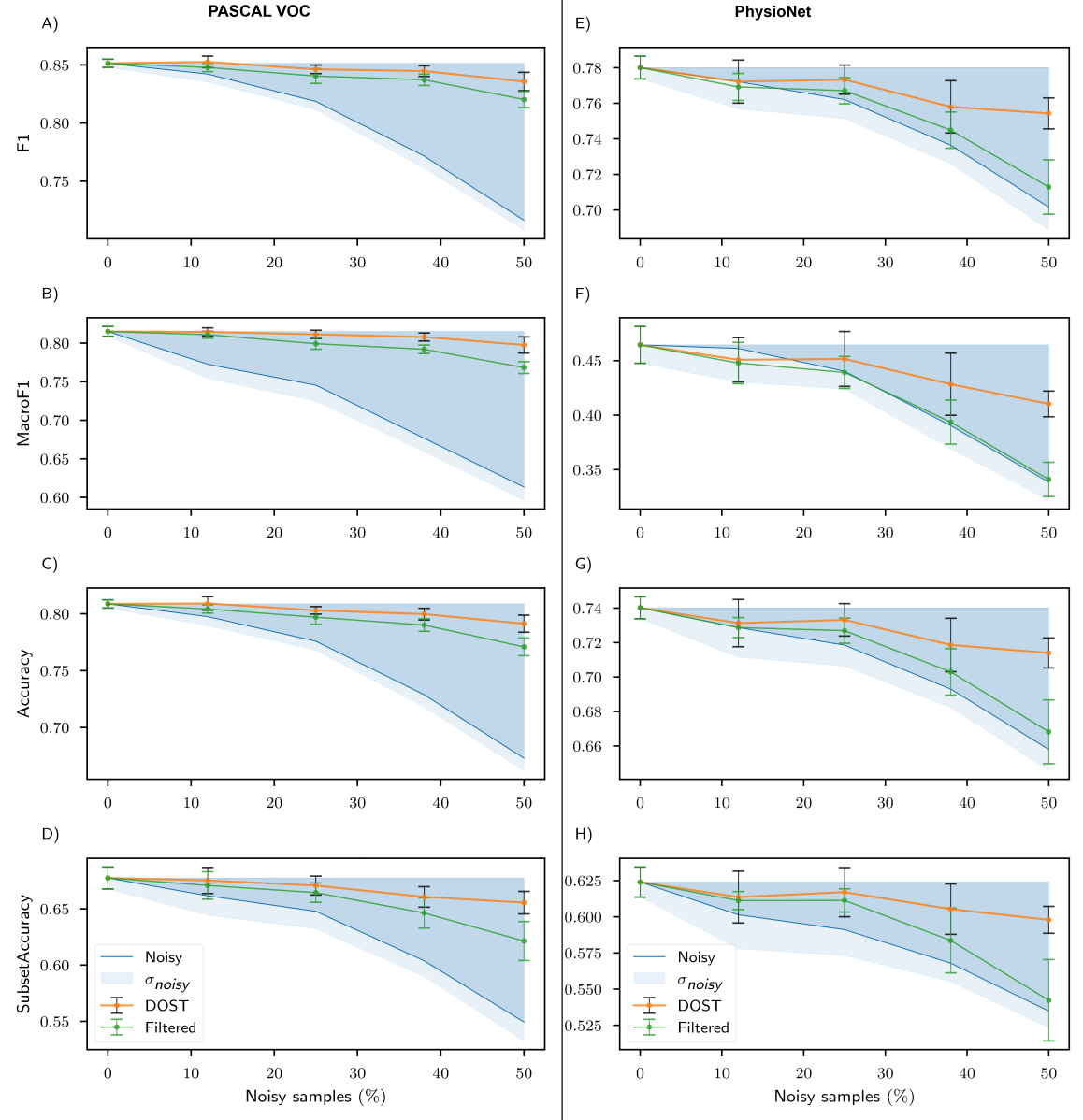}
	\label{fig:dost-effect}
\end{figure*}

Finally, we measure the effect on performance for the two strategies.
The \textbf{DOST} paradigm outperforms the naive filtering approach and shows significant improvement in F1, accuracy, subset accuracy, etc across the board (see Fig. \ref{fig:dost-effect}).
In particular, with the PASCAL VOC dataset, we can see that the effect of noise is almost eliminated by the DOST paradigm (see Table \ref{tab:pascal-dost}).

\begin{table}[!t]
	\centering
	\caption{Even with 50\% of instances containing noisy labels, DOST successfully counteracts the effect of noise.}
	\label{tab:pascal-dost}
	\begin{tabular}{|l|p{3em}|p{3em}|p{3em}|p{3em}|}
		\thickline
		Metric	& Perfect Dataset	& \textbf{DOST}	& Noisy Dataset	& Filtered Dataset	\\
		\thickline
		Noise Level	& None			& $50\%$	    & $50\%$		& $0\%$			\\
		\thickline
		Accuracy	& $0.809$ 	& $0.791$	& $0.673$ 	& $0.771$ 	\\
				& $\pm 0.004$	& $\pm 0.008$	& $\pm 0.011$	& $\pm 0.008$	\\
		\hline
		F1		& $0.851$ 	& $0.836$  	& $0.716$ 	& $0.820$ 	\\
				& $\pm 0.004$	& $\pm 0.008$	& $\pm 0.008$	& $\pm 0.007$	\\
		\hline
		Macro F1	& $0.815$ 	& $0.798$	& $0.613$ 	& $0.768$ 	\\
				& $\pm 0.007$	& $\pm 0.011$	& $\pm 0.017$	& $\pm 0.007$	\\
		\hline
		Subset Accuracy	& $0.677$ 	& $0.656$	& $0.549$ 	& $0.621$ 	\\
				& $\pm 0.010$	& $\pm 0.010$	& $\pm 0.017$	& $\pm 0.017$	\\
		\thickline
	\end{tabular}
\end{table}

The emperical study shows that the DOST paradigm not only reduces rule violations, thus making the models more aligned to domain rules, but it also showed performance gains.
This is achieved with repurposing data that would have otherwise been eliminated as noisy.
Since, all other factors are unchanged amongst the different training strategies, we conclude that the performance gains are likely due to inculcation of domain knowledge into the models.

%% file: chapters/conclusions.latex
\section{Conclusions}
\label{sec:conclusions}
The MLC problem arises in several practical scenarios and deep learning techniques have been widely used in this regard.
However, since large datasets are required to train such systems, the available datasets often contain significant quantities of label noise.
It is an ongoing problem in deep learning, and several strides have been made in this regard, but the problem hasn't received much attention when it comes to the MLC setting where such noise is more pronounced and complex.
The domains which would see the most benefit from wide adoption of deep learning techniques are also unfortunately the ones where this problem is especially pressing owing to annotation being cost prohibitive.
Additionally, deep learning systems do not natively abide by domain rules, and incorporating expert judgement and established knowledge into neural networks aid their deployability.

Our proposed \textbf{D}omain \textbf{O}bedient \textbf{S}elf-supervised \textbf{T}raining (\textbf{DOST}) scheme incorporates domain rules to minimize the negative impact of noisy annotations.
Instead of eliminating the noisy label instances (which is the conventional practice), \textbf{DOST} uses them for self-supervised training.
We demonstrate that this novel scheme not only shows improvements in domain obedience (aligning inferences with domain rules), but also has performance benefits in other relevant metrics.
We hope this work shows a path forward for further enquiries into assimilating expert knowledge into deep learning systems, thus expanding their applicability to critical scenarios.

%% file: chapters/appendix.latex
\clearpage
\newpage
\section{Appendix}
\label{sec:appendix}

The rules used in our experiments for the PhysioNet dataset \cite{cmpaper} were created in consultation with expert cardiologists based on well established facts about cardiovascular physiology. The rules used are as follows.

\begin{enumerate}
\item $IAVB \rightarrow \neg(AF \vee NSR)$
\item $AF \rightarrow \neg(IAVB \vee PR \vee SVPB \vee LPR \vee SA \vee SB \vee NSR \vee STach)$
\item $AFL \rightarrow \neg(PR \vee SVPB \vee LPR \vee SA \vee SB \vee NSR \vee STach)$
\item $Brady \rightarrow \neg(NSR \vee STach)$
\item $RBBB \rightarrow \neg(PR \vee NSR)$
\item $IRBBB \rightarrow \neg(PR \vee NSR)$
\item $LAnFB \rightarrow \neg(PR \vee NSR)$
\item $LBBB \rightarrow \neg(NSR)$
\item $NSIVCB \rightarrow \neg(NSR)$
\item $PR \rightarrow \neg(AF \vee AFL \vee RBBB \vee IRBBB \vee LAnFB \vee SA \vee SB \vee STach)$
\item $SVPB \rightarrow \neg(AF \vee AFL \vee NSR)$
\item $VPB \rightarrow \neg(NSR)$
\item $LPR \rightarrow \neg(AF \vee AFL \vee NSR)$
\item $LQT \rightarrow \neg(NSR)$
\item $QAb \rightarrow \neg(NSR)$
\item $RAD \rightarrow \neg(NSR)$
\item $SA \rightarrow \neg(AF \vee AFL \vee PR \vee NSR)$
\item $SB \rightarrow \neg(AF \vee AFL \vee PR \vee NSR \vee STach)$
\item $NSR \rightarrow \neg(IAVB \vee AF \vee AFL \vee Brady \vee RBBB \vee IRBBB \vee LAnFB \vee LBBB \vee NSIVCB \vee SVPB \vee VPB \vee LPR \vee LQT \vee QAb \vee RAD \vee SA \vee SB \vee STach \vee TAb \vee TInv)$
\item $STach \rightarrow \neg(AF \vee AFL \vee Brady \vee PR \vee SB \vee NSR)$
\item $TAb \rightarrow \neg(NSR)$
\item $TInv \rightarrow \neg(NSR)$
\end{enumerate}

Note that $\sim 20\%$ of annotated instances violated these rules in the PhysioNet dataset.

For the PASCAL VOC \cite{Everingham15} dataset, our rules are constructed from pairs that never occur together in the training/test set. These rules are constructed for demonstration purposes only, and may not hold when deployed in practice. The rules used are as follows.
\begin{enumerate}
\item $chair \rightarrow \neg(bus \vee dog \vee train \vee pottedplant \vee sofa \vee bird \vee cat \vee tvmonitor \vee motorbike \vee person \vee sheep \vee bottle \vee bicycle \vee aeroplane \vee horse \vee diningtable \vee car \vee boat \vee cow)$
\item $bus \rightarrow \neg(chair \vee sofa \vee bird \vee cat \vee tvmonitor \vee horse \vee diningtable \vee cow)$
\item $dog \rightarrow \neg(chair \vee train \vee aeroplane)$
\item $train \rightarrow \neg(chair \vee dog \vee bird \vee cat \vee tvmonitor \vee motorbike \vee sheep \vee bottle \vee aeroplane \vee diningtable \vee cow)$
\item $pottedplant \rightarrow \neg(chair \vee sheep \vee cow)$
\item $sofa \rightarrow \neg(chair \vee bus \vee bird \vee sheep \vee aeroplane \vee horse \vee cow)$
\item $bird \rightarrow \neg(chair \vee bus \vee train \vee sofa \vee motorbike)$
\item $cat \rightarrow \neg(chair \vee bus \vee train \vee motorbike \vee aeroplane \vee horse \vee boat)$
\item $tvmonitor \rightarrow \neg(chair \vee bus \vee train \vee motorbike \vee sheep \vee aeroplane \vee cow)$
\item $motorbike \rightarrow \neg(chair \vee train \vee bird \vee cat \vee tvmonitor \vee diningtable)$
\item $person \rightarrow \neg(chair)$
\item $sheep \rightarrow \neg(chair \vee train \vee pottedplant \vee sofa \vee tvmonitor \vee bicycle \vee aeroplane \vee diningtable \vee boat)$
\item $bottle \rightarrow \neg(chair \vee train \vee aeroplane \vee cow)$
\item $bicycle \rightarrow \neg(chair \vee sheep \vee aeroplane \vee horse)$
\item $aeroplane \rightarrow \neg(chair \vee dog \vee train \vee sofa \vee cat \vee tvmonitor \vee sheep \vee bottle \vee bicycle \vee horse \vee diningtable \vee cow)$
\item $horse \rightarrow \neg(chair \vee bus \vee sofa \vee cat \vee bicycle \vee aeroplane \vee diningtable \vee boat)$
\item $diningtable \rightarrow \neg(chair \vee bus \vee train \vee motorbike \vee sheep \vee aeroplane \vee horse \vee cow)$
\item $car \rightarrow \neg(chair)$
\item $boat \rightarrow \neg(chair \vee cat \vee sheep \vee horse)$
\item $cow \rightarrow \neg(chair \vee bus \vee train \vee pottedplant \vee sofa \vee tvmonitor \vee bottle \vee aeroplane \vee diningtable)$
\end{enumerate}